\documentclass[sigconf]{acmart}

\usepackage{soul}
\usepackage{algorithm}
\usepackage{algorithmic}
\usepackage{multirow}

\AtBeginDocument{%
  }

\copyrightyear{2026}
\acmYear{2026}
\setcopyright{cc}
\setcctype{by}
\acmConference[MM '26]{Proceedings of the 34th ACM International Conference on Multimedia}{November 10--14, 2026}{Rio de Janeiro, Brazil}
\acmBooktitle{Proceedings of the 34th ACM International Conference on Multimedia (MM '26), November 10--14, 2026, Rio de Janeiro, Brazil}
\acmDOI{10.1145/3767308.3835772}
\acmISBN{979-8-4007-2213-4/2026/11}

\begin{document}

\title[UG-UMRE]{UG-UMRE: Uncertainty-Guided Modality Augmentation and Distributional Calibration for Unified Multimodal Relation Extraction}

\author{Bo Kong}
\affiliation{%
  \institution{Xinjiang University}
  \city{Urumqi}
  \country{China}}
\email{107556522202@stu.xju.edu.cn}
\orcid{0000-0002-6002-9150}

\author{Liruizhi Jia}
\affiliation{%
  \institution{Xinjiang University}
  \city{Urumqi}
  \country{China}}
\email{jlrz@stu.xju.edu.cn}

\author{Yi Liang}
\affiliation{%
  \institution{Xinjiang University}
  \city{Urumqi}
  \country{China}}
\email{liangyi980826@hotmail.com}

\author{Chao Liu}
\affiliation{%
  \institution{Xinjiang University}
  \city{Urumqi}
  \country{China}}
\email{107556522208@stu.xju.edu.cn}

\author{Dongfang Han}
\affiliation{%
  \institution{Xinjiang University}
  \city{Urumqi}
  \country{China}}
\email{easth@stu.xju.edu.cn}

\author{Tianwei Yan}
\affiliation{%
  \institution{Chongqing Jiaotong University}
  \city{Chongqing}
  \country{China}}
\email{augusyan@cqjtu.edu.cn}

\author{Yuan Liu}
\affiliation{%
  \institution{Xinjiang University}
  \city{Urumqi}
  \country{China}}
\email{lyxju@xju.edu.cn}

\author{Shengquan Liu}
\authornote{Corresponding author}
\affiliation{%
  \institution{Xinjiang University}
  \city{Urumqi}
  \country{China}}
\email{liu@xju.edu.cn}


\renewcommand{\shortauthors}{Kong et al.}

\begin{abstract}
Unified Multimodal Relation Extraction (UMRE) aims to identify intra-modal and cross-modal relations between textual entities and visual objects. However, existing UMRE studies still encounter two critical issues: ignoring inherent aleatoric uncertainty causes noise propagation, and deep-seated heterogeneity between distinct modal distributions hinders alignment. To address these issues, we propose the Uncertainty-Guided UMRE Network (UG-UMRE). Specifically, we design an Uncertainty-Driven Unimodal Augmentation (UDUA) module, which models features as Gaussian distributions based on the Variational Information Bottleneck. By incorporating an uncertainty-aware self-supervised contrastive learning mechanism, UDUA effectively filters out noise while maintaining semantic consistency. Furthermore, we introduce the Joint Aleatoric Uncertainty Alignment (JAUA) module as a global semantic pre-calibration mechanism. JAUA leverages probabilistic distribution consistency to construct a shared latent space, eliminating the distributional gap by synchronizing cross-modal statistical properties, thereby laying a robust foundation for fine-grained interaction. Experiments on three benchmark datasets (UMRE, MORE, and MNRE) demonstrate that UG-UMRE achieves state-of-the-art performance. Further analysis validates the pluggable and effective performance of the proposed UDUA and JAUA modules.
\end{abstract}

\ccsdesc[500]{Information systems~Information extraction}
\ccsdesc[300]{Computing methodologies~Information extraction}

\keywords{Unified Multimodal Relation Extraction; Uncertainty-Guided Learning; Information Bottleneck; Modal Noise Processing; Multimodal Learning}

\maketitle

\section{Introduction}

With the ubiquity of multimodal learning and deep learning, research capabilities relying solely on text have become increasingly limited. Consequently, Multimodal Relation Extraction (MRE) has emerged as a pivotal research direction. Dedicated to extracting relation triplets using both textual and visual information, MRE plays an irreplaceable role in downstream applications such as multimodal knowledge graph construction, biomedical interaction modeling~\cite{b52}, visual question answering, and cross-modal retrieval~\cite{b39,b40}.

\begin{figure}[t]
    \centering
    \includegraphics[width=1.0\linewidth]{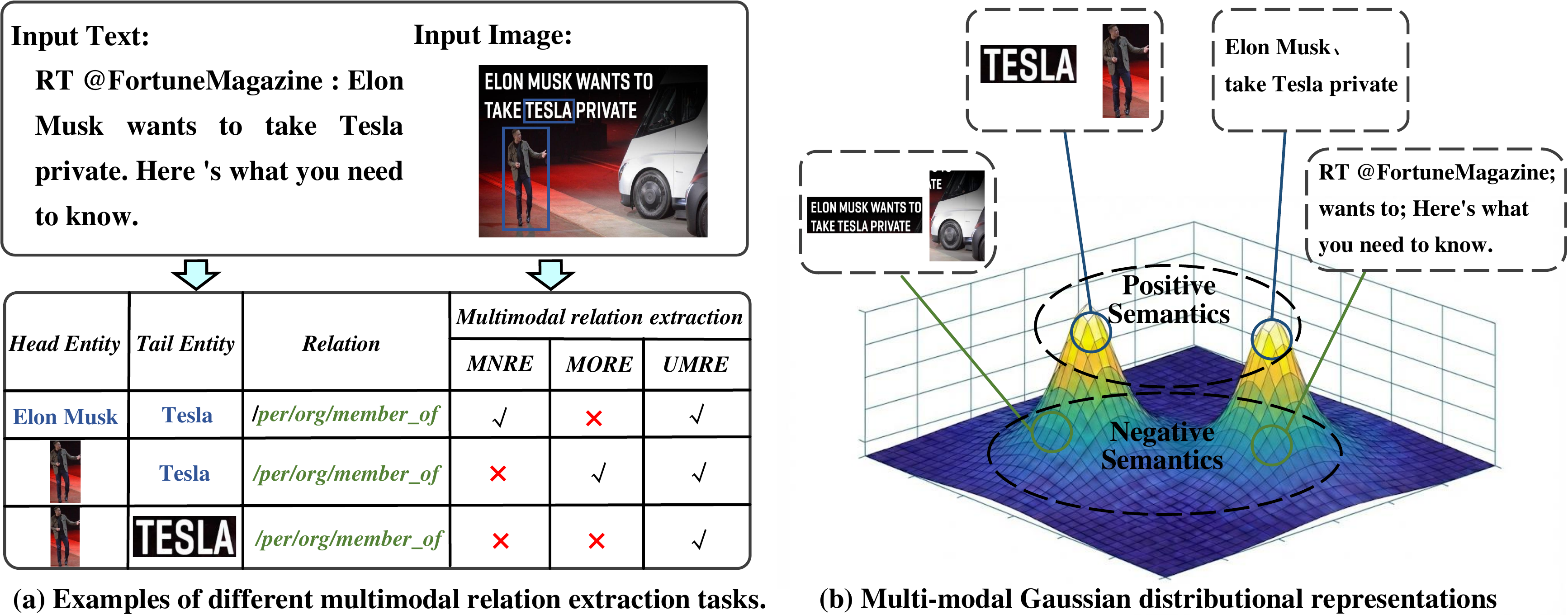}
    \caption{(a) Examples of different multimodal relation extraction tasks. Symbols \checkmark and $\times$ indicate whether the task successfully extracts the relational triplet or not. (b) Multi-modal Gaussian distributional representations illustrating fuzzy semantics in noisy latent space.}
    \label{fig:intro}
\end{figure}

As illustrated in Figure~\ref{fig:intro}(a), driven by the continuous expansion of the relation modeling scope, existing MRE research has evolved through three progressive subtasks. Early studies primarily focused on \textbf{Multimodal Neural Relation Extraction (MNRE)}~\cite{b2,b3,b4,b5,b7,b8,b9,b10,b11,b12,b13,b14}, which utilizes images merely as auxiliary clues to extract relations between entities within the textual modality. Subsequently, \textbf{Multimodal Object-Entity Relation Extraction (MORE)}~\cite{b6} was proposed to incorporate visual objects into relational triplets, focusing on mining cross-modal relations between textual entities and visual objects. However, multimodal interactions in real-world scenarios are inherently more intricate, making it difficult for these isolated task settings to satisfy the requirements for comprehensive data understanding. Consequently, \textbf{Unified Multimodal Relation Extraction (UMRE)}~\cite{b1} emerged. It breaks the boundaries of individual tasks, enabling the simultaneous extraction of both intra-modal and cross-modal relations between textual entities and visual objects within a unified framework.

Although UMRE provides a more comprehensive relational modeling paradigm that closely aligns with real-world scenarios, this high degree of unification drastically increases the complexity of cross-modal representation learning and semantic reasoning. Consequently, existing MRE methods still confront two critical scientific challenges when tackling the UMRE task:

\textbf{(1) Modal Noise}: Input data from different modalities suffer from inherent aleatoric uncertainty within their respective modal spaces. For instance, social media images often contain irrelevant background clutter, while textual descriptions may use ambiguous slang or abbreviations. Ignoring this inherent randomness leads to the erroneous propagation of noise, causing the model to overfit spurious correlations.
    
\textbf{(2) Modal Gap}: Since textual and visual representations originate from heterogeneous encoders, they reside in distinct feature spaces and distributions. This discrepancy hinders mutual semantic understanding between text and images; consequently, feeding uncalibrated features directly into fine-grained interaction inevitably leads to the failure of subsequent cross-modal fusion.

Existing methods typically perform cross-modal interactions directly on unfiltered and uncalibrated features, causing complex backend fusion networks to fail under the interference of modal noise and the distributional gap. To overcome this fundamental bottleneck, we propose the \textbf{Uncertainty-Guided UMRE Network (UG-UMRE)}. Our core idea is to model features as probabilistic distributions, strictly adhering to a new paradigm of ``denoising first, calibrating second, and fusing last'':

\textbf{Input Stage (Denoising \& Augmentation)}: Inspired by prior work~\cite{b5,b15,b16}, we propose the Uncertainty-Driven Unimodal Augmentation (UDUA) module. As illustrated in Figure~\ref{fig:intro}(b), each sample is represented by a multivariate Gaussian distribution, where the mean represents the stable core (corresponding to positive semantics), while the variance is explicitly quantified as the sample's aleatoric uncertainty (corresponding to negative semantics). To effectively leverage this aleatoric uncertainty, we design a reparamete-rization-based self-supervised contrastive learning mechanism: by conducting sampling within the variance-bounded semantic space, this mechanism forces the model to aggregate all potential semantic information toward the stable mean center, thereby achieving precise, distribution-aware denoising.

\textbf{Alignment Stage (Distributional Calibration)}: To mitigate the distributional gap between textual and visual representations induced by heterogeneous encoders, we design the Joint Aleatoric Uncertainty Alignment (JAUA) module. Existing methods typically perform cross-modal interaction directly on uncalibrated features, which is highly susceptible to semantic misalignment and fusion failure. In contrast, the JAUA module advocates for a global calibration of cross-modal features at the distribution level prior to deep interaction. Built upon the uncertainty-aware representations learned by the preceding UDUA module, this calibration step effectively eliminates cross-modal mismatches and constructs a highly consistent shared semantic space, thereby laying a solid foundation for subsequent reliable fine-grained feature fusion.

Extensive experiments on three public datasets (UMRE, MORE, and MNRE), which cover intra-modal and inter-modal semantic relations between textual entities and visual objects, demonstrate that UG-UMRE not only outperforms previous state-of-the-art baselines but also possesses excellent plug-and-play capabilities. The contributions of this paper are summarized as follows:
\begin{itemize}
    \item We propose the \textbf{UG-UMRE Network}. To the best of our knowledge, it pioneers the integration of uncertainty modeling into the UMRE task, providing a novel probabilistic solution to modal noise and distributional heterogeneity.
    \item We design the \textbf{UDUA} module, which achieves precise filtration of inherent unimodal noise and feature augmentation through variational Gaussian modeling and self-supervised contrastive learning.
    \item We propose the \textbf{JAUA} module, which effectively mitigates deep-seated heterogeneity and achieves fine-grained semantic calibration alignment by synchronizing cross-modal probabilistic distributions.
\end{itemize}

\section{Related Work}

\subsection{Unified Multimodal Relation Extraction}

MRE methods seek fine-grained cross-modal alignment while reducing modal noise. For MNRE, MEGA~\cite{b14} uses scene graphs to locate text-relevant regions; MKGformer~\cite{b9}, HVFormer~\cite{b4}, and IFAformer~\cite{b8} exploit hierarchical context and cross-modal attention; CGI-MRE~\cite{b3} and FocalMRE~\cite{b2} further suppress irrelevant information through noise filtering and refined attention. Related MRE studies emphasize structured interaction and augmented supervision. CSMA-CNER~\cite{b49} combines cross- and self-modal attention, MINIGE-MNER~\cite{b50} performs multi-stage interaction, and REIA~\cite{b51} jointly learns interaction policies and data augmentation. Efficient vision-language instruction tuning has also explored multi-head adapters and chain-of-thought reasoning~\cite{b53}. MOREformer~\cite{b6} introduces MORE and combines spatial and depth cues for object disambiguation, whereas REMOTE~\cite{b1} introduces UMRE and uses Multilevel Optimal Transport and Mixture-of-Experts mechanisms to model intra- and cross-modal relations.

Despite their effectiveness, including REMOTE, these methods generally lack denoising and distributional calibration before cross-modal interaction. Consequently, heterogeneous features from encoders such as BERT and ViT can be misaligned, while intra-modal aleatoric noise may be amplified during fusion.

\subsection{Uncertainty and Information Bottleneck}

Uncertainty in deep learning is commonly categorized as epistemic uncertainty in model parameters and aleatoric uncertainty inherent in data~\cite{b15}. The Information Bottleneck (IB)~\cite{b20} and Variational Information Bottleneck (VIB)~\cite{b21} learn compact representations that retain task-relevant information. In MRE, MRE-ISE~\cite{b7} applies Graph IB, and MMIB~\cite{b5} employs VIB for feature compression and alignment. However, many uncertainty-aware methods use uncertainty mainly for noise estimation or sample weighting~\cite{b16,b17,b18,b19}. Another line of work models uncertainty and dependencies in the supervision space, including label distribution learning~\cite{b42}, biased and instance-dependent noisy labels~\cite{b43,b48}, label-rank consistency for semi-supervised learning~\cite{b44}, heterogeneous label correlations and annotation quality in federated learning~\cite{b45,b47}, and positive-unlabeled distillation~\cite{b46}. These methods focus on label ambiguity, correlation, or incompleteness rather than input-level uncertainty.

In contrast, UG-UMRE models unimodal features as probabilistic distributions by coupling aleatoric uncertainty with VIB. UDUA uses the resulting uncertainty to drive targeted unimodal denoising, while JAUA exploits distributional properties to globally calibrate heterogeneous text and visual representations before fine-grained interaction. This design reduces both input noise and cross-modal distribution heterogeneity, providing more robust features for multimodal fusion.
\section{Our Proposed Method}

\begin{figure*}[t]
    \centering
    \includegraphics[width=0.8\linewidth]{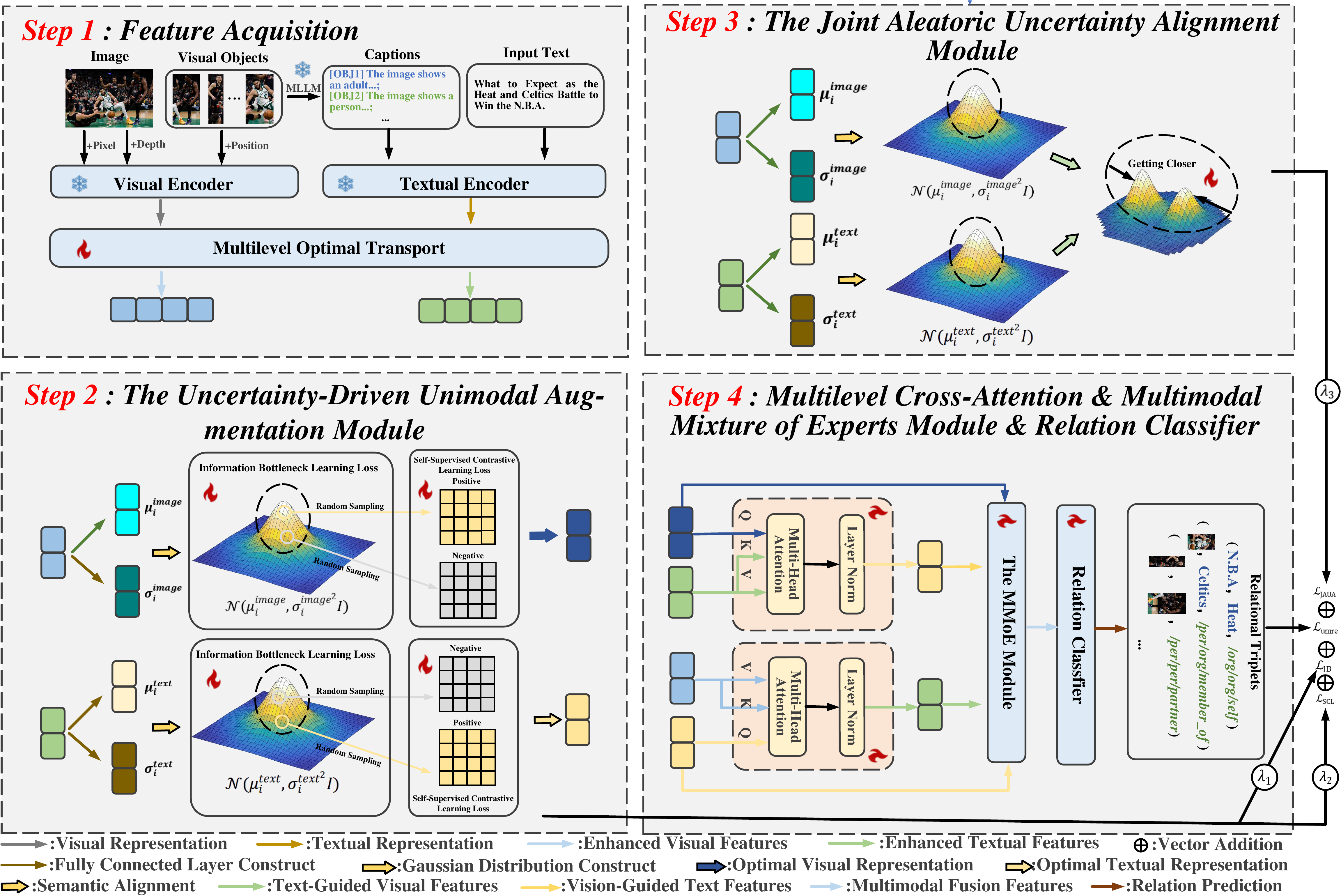}
    \caption{The overall framework of UG-UMRE.}
    \label{fig:framework}
\end{figure*}

In this section, we propose a novel UG-UMRE Network. As shown in Figure~\ref{fig:framework}, the framework consists of three key stages: First, intra-modal aleatoric noise is filtered via UDUA; second, \textbf{fine-grained semantic calibration alignment} is performed using JAUA to eliminate the distributional gap caused by heterogeneous encoders; finally, precise interaction and relation reasoning are achieved via MMoE in the calibrated feature space.

\subsection{Task Definition}
Before detailing the method, we first define the UMRE task. Specifically, given text $T$ and the corresponding image $I$, the UMRE task aims to extract entity relation triplets from the text with the aid of the image. The output results can be denoted as $\{(E_1, E_2, R_{12})_i\}_{i=1}^C$, where $E_1, E_2 \in \{ \mathcal{E}_{set}, \mathcal{O}_{set} \}$. Here, $(E_1, E_2, R_{12})_i$ represents the $i$-th relation triplet, $E_1, E_2$ denote two annotated entities, $\mathcal{E}_{set}$ is the set of textual entities, and $\mathcal{O}_{set}$ is the set of visual object entities. $R_{12}$ is the predefined relation between the entity pair, consisting of three different types: (1) intra-modal relations between textual entities $(e, e, r)$; (2) intra-modal relations between visual objects $(o, o, r)$; and (3) inter-modal relations between textual entities and visual objects $(e, o, r)$.

\subsection{Feature Acquisition}
\textbf{Textual Features.} Following REMOTE~\cite{b1}, we utilize Qwen2.5-VL-7B~\cite{b22} to generate captions and concatenate them with the text. The sequence is fed into BERT~\cite{b23} to obtain textual features $F_T \in \mathbb{R}^{n \times d_T}$, where $n$ is sequence length and $d_T$ is the hidden dimension.

\textbf{Visual Features.} Following REMOTE~\cite{b1}, we employ ViT~\cite{b24} and Depth-Anything~\cite{b25}. We resize object images to $H \times W$ to generate patch embeddings $E_{RGB}, E_{DEPTH} \in \mathbb{R}^{u \times d_V}$ ($u=\frac{H \times W}{P^2}$). Concatenated with position embeddings, we obtain visual features $F_V \in \mathbb{R}^{2u \times d_V}$.

\textbf{Multilevel Optimal Transport (MOT).} To mitigate information loss in sequential encoding~\cite{b1,b26,b27}, we adopt MOT. Leveraging Optimal Transport theory~\cite{b28,b29} and Sinkhorn optimization~\cite{b30}, it aggregates hierarchical representations to yield enhanced features $F_T' \in \mathbb{R}^{n \times d_T}$ and $F_V' \in \mathbb{R}^{2u \times d_V}$.

\subsection{Uncertainty-Driven Unimodal Augmentation Module}

Since data from different modalities contain unique noise in their respective feature spaces, traditional deterministic point embeddings cannot effectively characterize the inherent ambiguity of semantics. To this end, based on the VIB theory, we propose the UDUA module to model features as probability distributions to quantify aleatoric uncertainty and achieve feature augmentation through distribution-aware contrastive learning.

\subsubsection{VIB-based Uncertainty Quantification}

As shown in Step 2 of Figure~\ref{fig:framework}, we utilize the MOT-enhanced feature sequences $F_T'$ and $F_V'$ at the \textbf{holistic granularity}. Specifically, we first perform \textbf{global average pooling} to obtain holistic vectors, which are then projected via fully connected layers into a shared latent dimension $d_z$. These projected vectors parameterize the posterior distributions $p(Z|T) \sim \mathcal{N}(\mu_T, \sigma_T^2I)$ and $p(Z|I) \sim \mathcal{N}(\mu_I, \sigma_I^2I)$, aiming to capture global semantic uncertainty (e.g., scene ambiguity) rather than local token noise.

To guide information retention, we introduce modality-specific target variables $U_T$ and $U_V$ (both set to ground-truth labels $Y$). The objective is to maximally retain information beneficial for these targets while filtering out redundant noise via the Information Bottleneck principle:
\begin{equation}
    \mathcal{L}_{IB} = \beta_1 I(Z_T; F_T') + \beta_2 I(Z_V; F_V') - I(Z_T; U_T) - I(Z_V; U_V)
\end{equation}

We optimize Eq. (2) using variational approximations. The information compression terms are minimized via KL divergence against a standard Gaussian prior $r(Z) \sim \mathcal{N}(0, I)$:
\begin{equation}
    I(Z_T; F_T') \approx \mathbb{D}_{KL}(p_{\theta_T}(Z_T|F_T') || r(Z_T))
\end{equation}

The information retention terms are converted into cross-entropy losses via variational approximations $q(U|Z)$:
\begin{equation}
    -I(Z_T; U_T) \approx \mathcal{L}_{CE}(q_{\theta_T}(U_T|Z_T), U_T)
\end{equation}

This formulation allows the model to effectively filter aleatoric noise through the bottleneck constraint while ensuring the learned features remain discriminative for the final relation extraction task.

\subsubsection{Uncertainty-Aware Self-Supervised Contrastive Learning}

Thr-ough VIB optimization, the obtained mean $\mu$ represents stable core semantics, while the learned variance $\sigma^2$ quantifies the aleatoric uncertainty. To utilize this uncertainty for augmentation without introducing meaningless noise, we introduce a reparameterization-based sampling strategy~\cite{b31}. Specifically, for the $i$-th sample, we use its mean $\mu_i$ as an anchor and sample a ``positive sample'' $z_i^+$ from its distribution:
\begin{equation}
    z_i^+ = \mu_i + \sigma_i \odot \epsilon
\end{equation}
Where $\epsilon \sim \mathcal{N}(0, I)$, and the perturbation is explicitly constrained by the learned standard deviation $\sigma_i$ (derived from the variance $\sigma_i^2$ in Eq. 4). The model automatically predicts small variances for semantically clear samples and large variances for ambiguous ones.

To force the model to maintain semantic consistency in the uncertain latent space, we design the following SCL loss $\mathcal{L}_{SCL}$ utilizing an InfoNCE-style objective:
\begin{equation}
\resizebox{.91\linewidth}{!}{$
    \mathcal{L}_{SCL} = - \log \frac{\exp(\text{sim}(\mu_i, z_i^+) / \tau)}{\exp(\text{sim}(\mu_i, z_i^+) / \tau) + \sum_{j \neq i}^N \exp(\text{sim}(\mu_i, z_j) / \tau)}
$}
\end{equation}
Where $z_i^+$ is the positive sample generated from the anchor $\mu_i$, and $\{z_j\}_{j \neq i}^N$ denotes the set of negative samples \textbf{sampled from the distributions of other instances} within the same mini-batch. $\tau$ is the temperature factor, and $\text{sim}$ is cosine similarity. In contrast to deterministic InfoNCE, the positive and negative views are sampled from uncertainty-bounded Gaussian neighborhoods. Thus, the loss regularizes semantic consistency over plausible feature perturbations rather than over a fixed point embedding, while the VIB objective constrains these perturbations to remain task-relevant. This pulls semantic variants toward the stable center $\mu_i$ while separating them from unrelated instances.

\subsection{Joint Aleatoric Uncertainty Alignment Module}

Although UDUA provides denoised unimodal features, the textual and visual representations remain in heterogeneous feature spaces. Directly feeding these uncalibrated features into interaction modules inevitably leads to semantic misalignment. To this end, we design the \textbf{JAUA module} for global semantic calibration. Unlike traditional unidirectional alignment, JAUA operates in the \textbf{distributional domain} to synchronize the statistical properties of heterogeneous modalities. This constructs a shared latent space that eliminates the distributional gap while preserving modality-specific details, laying a reliable foundation for subsequent fine-grained interaction. Formally, the bidirectional KL divergence is calculated as follows:

\begin{equation}
\resizebox{.91\linewidth}{!}{$
    \mathcal{L}_{JAUA} = \frac{1}{2} (\mathbb{D}_{KL}(p(Z_T|F_T') || p(Z_V|F_V')) + \mathbb{D}_{KL}(p(Z_V|F_V') || p(Z_T|F_T')))
$}
\label{eq:jaua}
\end{equation}

For the diagonal Gaussian posteriors produced by UDUA, the KL term has a closed form. Let $q_T=\mathcal{N}(\mu_T,\mathrm{diag}(\sigma_T^2))$ and $q_V=\mathcal{N}(\mu_V,\mathrm{diag}(\sigma_V^2))$. Then
\begin{equation}
\resizebox{.91\linewidth}{!}{$
\mathbb{D}_{KL}(q_T\Vert q_V)=\frac{1}{2}\sum_{k=1}^{d_z}\left[\log\frac{\sigma_{V,k}^{2}}{\sigma_{T,k}^{2}}+\frac{\sigma_{T,k}^{2}+(\mu_{T,k}-\mu_{V,k})^{2}}{\sigma_{V,k}^{2}}-1\right].
$}
\end{equation}

The symmetric objective therefore jointly penalizes disagreement between semantic centers and disagreement between uncertainty scales. This distribution-level calibration acts only on holistic latent variables. The multi-level sequences $F_T'$ and $F_V'$ are retained for the subsequent interaction module, preserving local modality-specific cues instead of overwriting them with a shared representation.

\subsection{Multimodal Mixture-of-Experts Module}

Benefiting from the shared latent space constructed by JAUA, textual and visual features are globally pre-calibrated. To capture fine-grained semantic associations without discarding local cues, we introduce a Hierarchical Cross-Modal Interaction (HCMI) method. Let $\bar F_A$ denote the highest-level feature of source modality $A$, and let $F_B^{(j)}$ denote the level-$j$ feature of the complementary modality $B$, where $(A,B)\in\{(V,T),(T,V)\}$. HCMI uses a high-level source query to attend to each target level through explicit linear projections:
\begin{equation}
\begin{aligned}
Q_j^{A\to B}&=\bar F_AW_{Q,j}^{A\to B}, & K_j^{A\to B}&=F_B^{(j)}W_{K,j}^{A\to B},\\
V_j^{A\to B}&=F_B^{(j)}W_{V,j}^{A\to B}, & F_j^{A\to B}&=\mathrm{Softmax}\left(\frac{Q_j^{A\to B}(K_j^{A\to B})^\top}{\sqrt{d}}\right)V_j^{A\to B}.
\end{aligned}
\end{equation}
Where $W_{Q,j}^{A\to B}$, $W_{K,j}^{A\to B}$, and $W_{V,j}^{A\to B}$ are learnable maps into a common $d$-dimensional attention space. Instantiating the equation for $(A,B)=(V,T)$ and $(T,V)$ produces the visual-guided textual features $F_j^{V\to T}$ and the text-guided visual features $F_j^{T\to V}$, respectively.

The MMoE module then dynamically assigns expert weights to optimal textual/visual representations and cross-modal interaction features based on relation triplets to obtain the final fused feature $H$:
\begin{equation}
    H = \text{MMoE}(Z_T, Z_V, F^{V \to T}, F^{T \to V})
\end{equation}

\subsection{Multimodal Relation Extraction}

We select the final representation based on the relation triplet type. For textual entities, we use the text feature separated by $\langle s \rangle$; for visual objects, we use the text feature of the caption separated by $\langle o \rangle$ combined with the visual feature:
\begin{equation}
    h = \begin{cases} H_{\langle s \rangle}^{T,V} & \text{if the target is a textual entity} \\ [H_{\langle o_i \rangle}^{T,V}, H_{\langle o_i \rangle}^{V,T}] & \text{if the target is a visual entity} \end{cases}
\end{equation}

The prediction is made via an MLP:
\begin{equation}
    P_r = \text{argmax}(\text{MLP}([h_{head}, h_{tail}]))
\end{equation}

We minimize the cross-entropy loss between the prediction and ground truth:
\begin{equation}
    \mathcal{L}_{umre} = - \sum y_r \log(P_r)
\end{equation}

\subsection{Joint Training}
The final objective combines the primary task with the fine-grained alignment optimization:
\begin{equation}
    \mathcal{L} = \mathcal{L}_{umre} + \lambda_1 \mathcal{L}_{IB} + \lambda_2 \mathcal{L}_{SCL} + \lambda_3 \mathcal{L}_{JAUA}
\end{equation}

Where $\lambda_1, \lambda_2, \lambda_3$ are hyperparameters balancing the contributions of unimodal IB loss, SCL loss, and JAUA alignment loss.

\begin{table}[t]
    \centering
    \caption{Detailed statistics of the MNRE, MORE, and UMRE datasets used in our experiments.}
    \label{tab:datasets}
    \resizebox{\columnwidth}{!}{
    \begin{tabular}{lccccc}
        \toprule
        \textbf{Dataset} & \textbf{Images} & \textbf{Sentences} & \textbf{\begin{tabular}[c]{@{}c@{}}Visual\\Objects\end{tabular}} & \textbf{Triplets} & \textbf{Relations} \\
        \midrule
        MNRE & 9,201 & 9,201 & - & 15,485 & 23 \\
        MORE & 3,559 & 3,559 & 13,520 & 20,264 & 21 \\
        UMRE & 12,737 & 12,737 & 20,978 & 55,021 & 28 \\
        \bottomrule
    \end{tabular}
    }
\end{table}

\begin{table*}[t]
    \centering
    \caption{Overall performance comparison on UMRE, MORE, and MNRE datasets. \textbf{Bold} indicates the best result, \underline{underline} indicates the second best. $\dag$ denotes results reproduced by us. "-" indicates missing values in baseline papers.}
    \label{tab:main_results}
    \begin{tabular}{lc ccc ccc ccc}
        \toprule
        \multirow{2}{*}{\textbf{Methods}} & \multirow{2}{*}{\textbf{Venue}} & \multicolumn{3}{c}{\textbf{UMRE}} & \multicolumn{3}{c}{\textbf{MORE}} & \multicolumn{3}{c}{\textbf{MNRE}} \\
        \cmidrule(lr){3-5} \cmidrule(lr){6-8} \cmidrule(lr){9-11}
         & & P (\%)& R (\%)& F1 (\%)& P (\%)& R (\%)& F1 (\%)& P (\%)& R (\%)& F1 (\%)\\
        \midrule
        Qwen2-VL-7B~\cite{b35} & arXiv24 & 7.32 & 11.52 & 8.95 & 8.52 & 12.87 & 10.25 & 11.42 & 13.09 & 12.20 \\
        Qwen2.5-VL-7B~\cite{b22} & arXiv25 & 12.33 & 12.04 & 12.19 & 25.81 & 23.95 & 24.84 & 13.82 & 15.63 & 14.68 \\
        \begin{tabular}[c]{@{}l@{}}Llama-3.2-11B-Vision~\cite{b36}\end{tabular} & arXiv24 & 5.98 & 13.46 & 8.28 & 5.09 & 18.66 & 8.01 & 9.89 & 13.32 & 11.35 \\
        \midrule
        MEGA~\cite{b14} & MM21 & 49.51 & 47.24 & 48.34 & 33.30 & 38.53 & 35.72 & 64.51 & 68.44 & 66.41 \\
        MKGFormer~\cite{b9} & SIGIR22 & 60.22 & 60.26 & 60.23 & 55.76 & 53.74 & 54.73 & 82.40 & 81.73 & 82.06 \\
        IFAformer~\cite{b8} & AAAI23 & 60.21 & 62.04 & 61.11 & 55.13 & 54.24 & 54.68 & 82.59 & 80.78 & 81.67 \\
        MRE-ISE~\cite{b7} & ACL23 & - & - & - & - & - & - & 84.69 & 83.38 & 84.03 \\
        MOREformer~\cite{b6} & MM23 & 63.03 & 62.77 & 62.89 & 62.18 & 63.34 & 62.75 & 82.19 & 82.35 & 82.27 \\
        MMIB~\cite{b5} & TASLP24 & 61.58 & 60.74 & 61.16 & 60.15 & 62.28 & 61.17 & 83.49 & 82.97 & 83.23 \\
        HVFormer~\cite{b4} & WWW24 & 61.37 & 61.04 & 61.20 & 58.81 & 62.84 & 60.76 & 84.14 & 82.65 & 83.39 \\
        CGI-MRE~\cite{b3} & ICMR24 & 60.45 & 61.84 & 61.63 & 57.44 & 63.01 & 60.09 & 85.02 & 84.22 & 84.62 \\
        FocalMRE$^\dag$~\cite{b2} & MM24 & 61.35 & 65.96 & 63.57 & 60.57 & 62.67 & 61.60 & 87.97 & 86.88 & 87.42 \\
        REMOTE$^\dag$~\cite{b1} & MM25 & 66.24 & 69.10 & 67.64 & 61.48 & 64.21 & 62.81 & \underline{88.16} & 87.26 & 87.70 \\
        \midrule
        FocalMRE+UDUA & - & 61.53 & 66.74 & 64.03 & 62.32 & 64.28 & 63.28 & 88.14 & 87.34 & 87.73 \\
        FocalMRE+JAUA & - & 62.27 & 67.01 & 64.01 & 62.83 & 63.81 & 63.62 & 88.60 & 87.18 & 87.88 \\
        \begin{tabular}[c]{@{}l@{}}FocalMRE+UDUA+JAUA\end{tabular} & - & 62.45 & 68.84 & 65.49 & 63.44 & 65.01 & 64.22 & 89.30 & 87.34 & 88.31 \\
        REMOTE+UDUA & - & 67.04 & \underline{70.50} & \underline{68.73} & 65.17 & \underline{65.84} & \underline{65.50} & \underline{89.78} & \underline{87.81} & \underline{88.78} \\
        REMOTE+JAUA & - & \underline{67.34} & 69.77 & 68.53 & \underline{65.19} & 65.32 & 65.25 & 89.62 & 87.66 & 88.63 \\
        \begin{tabular}[c]{@{}l@{}}REMOTE+UDUA+JAUA\end{tabular} & - & \textbf{69.16} & \textbf{70.82} & \textbf{69.98} & \textbf{66.79} & \textbf{66.72} & \textbf{66.76} & \textbf{90.44} & \textbf{88.75} & \textbf{89.59} \\
        \bottomrule
    \end{tabular}
\end{table*}

\section{Experiments}

\subsection{Experimental Setup}
\textbf{Datasets \& Metrics.} We evaluate on MNRE~\cite{b34}, MORE~\cite{b6}, and UMRE~\cite{b1} (statistics in Table~\ref{tab:datasets}). We report Precision (P), Recall (R), and F1-score (F1).

\textbf{Implementation Details.} All experiments were conducted on a single NVIDIA RTX 4090 GPU, utilizing BERT-base and ViT-B/32 as the backbone encoders. Feature dimensions are 768 (text) and 4096 (visual objects). The models were optimized using the AdamW optimizer~\cite{b33} for 30 epochs (batch size 16, lr 1e-5). Hyperparameters: $\beta_1 = \beta_2 = 1$, $\tau = 0.175$, $\lambda_1 = 1e-3$, $\lambda_2 = 1e-5$, $\lambda_3 = 1e-3$.

\subsection{Main Results}

\textbf{Baselines.} We compare UG-UMRE with two categories of state-of-the-art methods:
(1) \textbf{MRE Models:} MEGA~\cite{b14}, MKGFormer~\cite{b9}, IFAformer~\cite{b8}, MRE-ISE~\cite{b7} MOREformer~\cite{b6}, MMIB~\cite{b5}, HVFormer~\cite{b4}, CGI-MRE~\cite{b3}, FocalMRE~\cite{b2}, and REMOTE~\cite{b1}.
(2) \textbf{MLLMs:} Qwen2-VL-7B~\cite{b35}, Qwen2.5-VL-7B~\cite{b22}, and Llama-3.2-11B-Vision~\cite{b36}.

\textbf{Fairness \& Adaptability Setup:} We evaluate the proposed modules through two controlled integrations (marked with $^\dag$). For the primary comparison, REMOTE and REMOTE+UG-UMRE use identical visual (Depth+RGB) and textual (Caption+Text) inputs, including the same Depth-Anything features and Qwen2.5-VL-generated captions. For the additional control, FocalMRE and FocalMRE+UG-UMRE use the same native FocalMRE inputs, without these added features. 

This differentiated setup verifies consistent gains across diverse input configurations without relying on a particular feature-engine-ering recipe.

Methods such as TMR~\cite{b10}, ES-MRE~\cite{b11}, HGMAF~\cite{b12}, and CAMIM~\cite{b13} adopt text-to-image generation~\cite{b37} for augmentation. However, this modifies the original layout and positional semantics, rendering it inappropriate for unified relation extraction which demands strict spatial preservation. Thus, methods involving generative augmentation are not included.

\begin{table}[t]
    \centering
        \caption{Ablation study on the UMRE, MORE, and MNRE datasets. ``w/o'' denotes the removal of a specific module. \textbf{Bold} indicates the best result.}
    \label{tab:ablation}
    \resizebox{\columnwidth}{!}{
    \begin{tabular}{lcccccc}
        \toprule
        \multirow{2}{*}{\textbf{Model}} & \multicolumn{2}{c}{\textbf{UMRE}} & \multicolumn{2}{c}{\textbf{MORE}} & \multicolumn{2}{c}{\textbf{MNRE}} \\
        \cmidrule(lr){2-3} \cmidrule(lr){4-5} \cmidrule(lr){6-7}
         & Acc (\%)& F1 (\%)& Acc (\%)& F1 (\%)& Acc (\%)& F1 (\%)\\
        \midrule
        \textbf{UG-UMRE} & \textbf{79.27} & \textbf{69.98} & \textbf{86.37} & \textbf{66.76} & \textbf{95.17} & \textbf{89.59} \\
        -w/o UDUA (Text) & 78.02 & 68.77 & 84.79 & 65.23 & 94.73 & 88.66 \\
        -w/o UDUA (Image) & 77.92 & 68.35 & 84.54 & 64.91 & 94.04 & 88.24 \\
        -w/o UDUA (SCL) & 78.64 & 69.33 & 85.33 & 65.91 & 94.89 & 88.97 \\
        \bottomrule
    \end{tabular}
    }
\end{table}

\textbf{Overall Performance.} We integrated UG-UMRE into the current SOTA model, REMOTE, for comprehensive evaluation. As shown in Table~\ref{tab:main_results}, our method outperforms all baselines on the UMRE, MORE, and MNRE datasets. Specifically, compared to the original REMOTE model, UG-UMRE improves the F1-score by \textbf{2.34\%}, \textbf{3.95\%}, and \textbf{2.25\%}, reaching new SOTA levels of \textbf{69.98\%}, \textbf{66.76\%}, and \textbf{89.59\%}, respectively. Based on the results, we make the following observations:

(1) \textbf{Limitations of General Multimodal Large Language Models (MLLMs)}: MLLMs (e.g., Qwen2.5-VL) generally perform poorly on the UMRE task. This indicates that despite their strong generalization, MLLMs struggle with fine-grained information extraction requiring precise semantic boundaries and complex relational reasoning, highlighting the need for specialized models.

(2) \textbf{Necessity of Global Calibration}: Methods like REMOTE outperform coarse-grained methods by enhancing unimodal details via MOT. However, their performance is still capped by the cross-modal distributional gap. The significant improvement of UG-UMRE proves that ``structural aggregation'' alone is insufficient; global semantic pre-calibration (JAUA) is essential to break the performance bottleneck caused by heterogeneity.

(3) \textbf{Synergy between Denoising and Calibration}: Results show that introducing UDUA or JAUA individually improves performance, but the full model performs best. Notably, UDUA yields slightly higher gains than JAUA alone, revealing a deep dependency: high-quality, denoised unimodal features (via UDUA) are a prerequisite for effective cross-modal alignment (via JAUA).

(4) \textbf{Plug-and-Play Capability}: Integrating UG-UMRE into both FocalMRE (standard inputs) and REMOTE (enhanced inputs) yields consistent improvements. Crucially, these gains are achieved \textbf{without altering the original input/output interfaces or core parameters} of the backbone networks. This demonstrates that UG-UMRE acts as a lightweight, non-intrusive ``safety valve'' and ``calibrator,'' providing a universal solution for mitigating modal noise and distributional heterogeneity regardless of the underlying model architecture or feature richness.

\subsection{Ablation Study}

We evaluate UDUA on the UMRE, MORE, and MNRE datasets (Table~\ref{tab:ablation}). Removing UDUA from either modality reduces performance, with a larger F1 decline for the visual branch (1.63\% on UMRE) than for the textual branch (1.21\%). This confirms that visual noise is more harmful, since objects in complex scenes have less explicit semantic boundaries than textual tokens. Without UDUA's probabilistic constraint, visual features are more susceptible to background clutter and can mislead cross-modal fusion.

Removing uncertainty-aware SCL also degrades performance. Although VIB filters noise and preserves task-relevant information, it does not explicitly enforce discriminability. SCL therefore anchors stable semantics in the noisy latent space, helping distinguish reliable information from noise.
\begin{figure}[t]
    \centering
    \includegraphics[width=1.0\columnwidth]{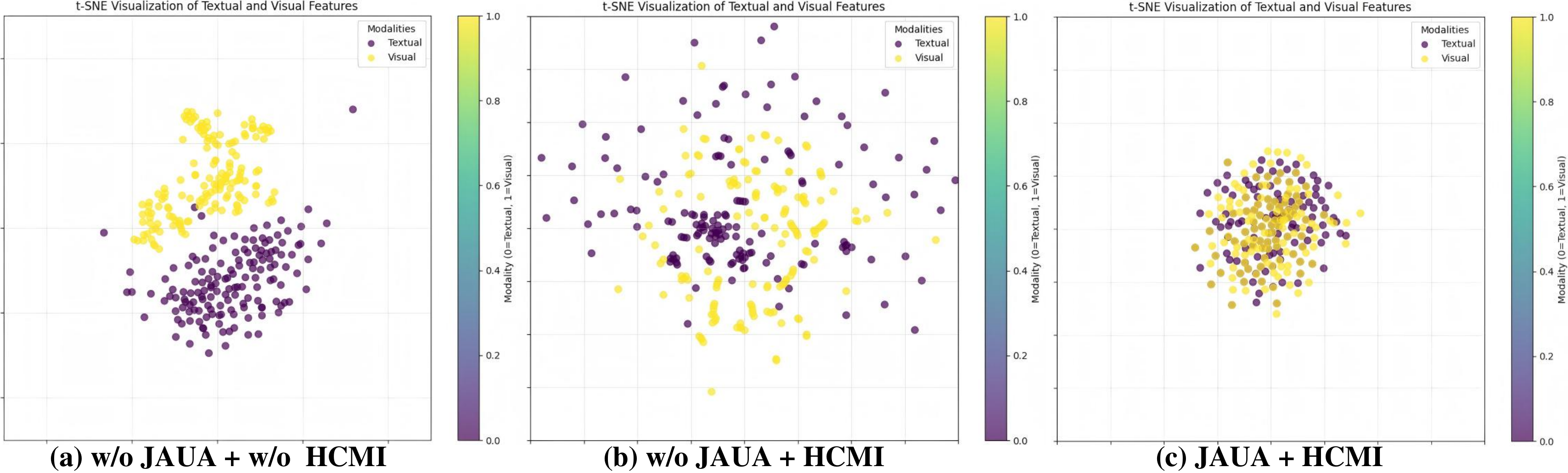}
    \caption{t-SNE visualization of feature distributions.}
    \label{fig:tsne}
\end{figure}
\subsection{Visualization of Modal Gap}

To validate the effectiveness of the JAUA module in eliminating cross-modal distributional heterogeneity, we conducted visualization experiments on 150 text-image pairs from the UMRE test set. We employed t-SNE~\cite{b38} (with 500 iterations and a perplexity of 10) to compare feature distributions under three settings: without JAUA and HCMI (w/o JAUA + w/o HCMI), with HCMI only (w/o JAUA + HCMI), and with both JAUA and HCMI (JAUA + HCMI), as shown in Figure~\ref{fig:tsne} (a), (b), and (c).

The results indicate that: (1) Initial features exhibit a significant distributional gap. (2) While relying solely on HCMI can reduce the modal distance, the distributions remain loosely separated, failing to fundamentally resolve the underlying heterogeneity. (3) In contrast, with the introduction of JAUA, textual and visual features become tightly interwoven and overlapped within the latent space. This demonstrates that JAUA achieves global semantic pre-calibration by synchronizing statistical properties, effectively eliminating the distributional gap and laying a solid alignment foundation for fine-grained interaction.

\subsection{Subtask Performance Analysis}

Table~\ref{tab:subtasks} compares the performance of SOTA models specifically trained on individual UMRE subtasks. The experimental results clearly show that both FocalMRE and REMOTE achieve significant improvements across all subtasks after integrating the UG-UMRE network. These results demonstrate that, even within specific triplet types, our proposed UDUA and JAUA modules are capable of effectively addressing the challenges of modal noise interference and cross-modal semantic heterogeneity inherent in the UMRE task.
\begin{table}[t]
    \centering
        \caption{Performance comparison on UMRE subtasks. \textbf{Bold} indicates the best result.}
    \label{tab:subtasks}
    \resizebox{\columnwidth}{!}{
    \begin{tabular}{lcccccc}
        \toprule
        \multirow{2}{*}{\textbf{Model}} & \multicolumn{2}{c}{\textbf{Entity-Entity}} & \multicolumn{2}{c}{\textbf{Object-Object}} & \multicolumn{2}{c}{\textbf{Entity-Object}} \\
        \cmidrule(lr){2-3} \cmidrule(lr){4-5} \cmidrule(lr){6-7}
         & Acc (\%)& F1 (\%)& Acc (\%)& F1 (\%)& Acc (\%)& F1 (\%)\\
        \midrule
        FocalMRE & 73.52 & 61.72 & 68.24 & 41.07 & 76.33 & 69.08 \\
        REMOTE & 73.81 & 62.46 & 67.77 & 44.86 & 80.74 & 72.79 \\
        FocalMRE+UG & 75.44 & 63.58 & 70.03 & 42.31 & 78.33 & 71.17 \\
        REMOTE+UG & \textbf{79.28} & \textbf{68.00} & \textbf{73.60} & \textbf{45.24} & \textbf{82.31} & \textbf{76.10} \\
        \bottomrule
    \end{tabular}
    }
\end{table}
\begin{table}[t]
    \centering
    \caption{Performance comparison on the Aug-Noise challenge set (F1-score(\%)) with varying noise ratios. \textbf{Bold} indicates the best result.}
    \label{tab:noise}
    \begin{tabular}{lccc}
        \toprule
        \textbf{Dataset} & \textbf{Ratio} & \textbf{REMOTE} & \textbf{REMOTE+UDUA} \\
        \midrule
    \multirow{4}{*}{Aug-Noise} & 5\%   & 66.03 & \textbf{68.27} (+2.24) \\
          & 20\%  & 61.55 & \textbf{66.83} (+5.28) \\
          & 50\%  & 48.06 & \textbf{62.52} (+14.46) \\
          & 100\% & 35.01 & \textbf{52.33} (+17.32) \\
    \midrule
    \multirow{4}{*}{Aug-Noise (text)} & 5\%   & 67.35 & \textbf{68.54} (+1.19) \\
          & 20\%  & 65.50 & \textbf{67.81} (+2.31) \\
          & 50\%  & 62.07 & \textbf{66.53} (+4.46) \\
          & 100\% & 56.12 &\textbf{63.14} (+7.02) \\
    \midrule
    \multirow{4}{*}{Aug-Noise (image)} & 5\%   & 66.85 & \textbf{68.43} (+1.58) \\
          & 20\%  & 63.54 & \textbf{67.52} (+3.98) \\
          & 50\%  & 55.02 & \textbf{65.30} (+10.28) \\
          & 100\% & 46.07 & \textbf{60.78} (+14.71) \\
        \bottomrule
    \end{tabular}
\end{table}


\begin{figure}[t]
    \centering
    \includegraphics[width=1.0\columnwidth]{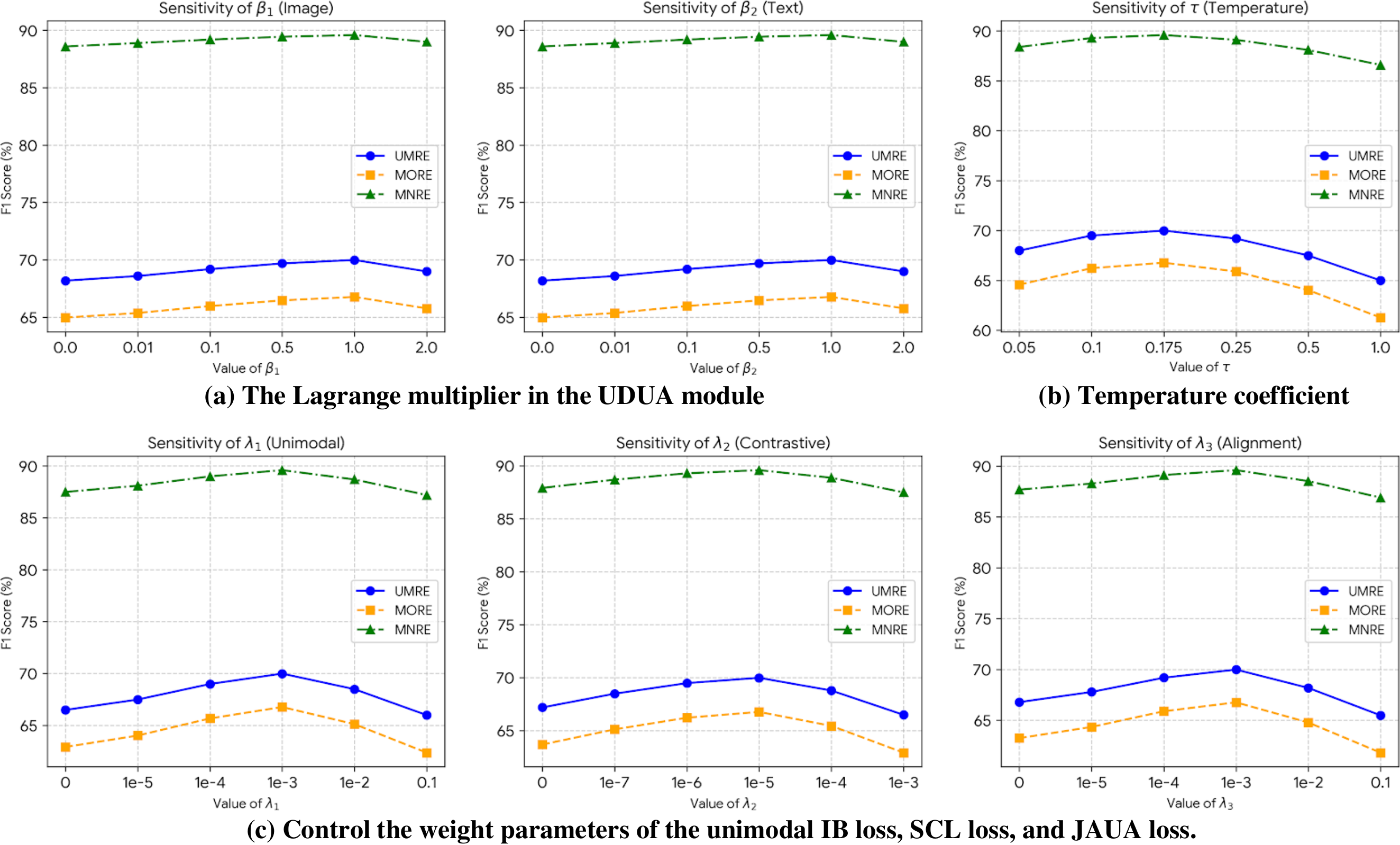}
    \caption{Sensitivity analysis of hyperparameters on the UMRE dataset. (a) Impact of Lagrange multipliers $\beta$. (b) Impact of temperature $\tau$. (c) Impact of loss weights $\lambda$.}
    \label{fig:params}
\end{figure}

\subsection{Noise Robustness Analysis}
Table~\ref{tab:noise} demonstrates the robustness of the UDUA module on the Aug-Noise challenge set. To construct this set, we introduced controllable aleatoric noise: for text, we applied random token masking with varying ratios; for images, we applied Gaussian noise perturbation to the visual features. Under extreme conditions (100\% noise), the baseline REMOTE suffers a catastrophic drop to 35.01\% F1, whereas REMOTE+UDUA maintains 52.33\%, achieving a significant gain of 17.32\%. Consistent improvements in visual (+14.71\%) and text (+7.02\%) scenarios further verify the "safety valve" mechanism of UDUA: by quantifying uncertainty, it dynamically down-weights high-variance noisy features, effectively blocking noise propagation and preventing performance collapse.

\subsection{Parameter Sensitivity}

To investigate the impact of hyperparameters on the performance of UG-UMRE (based on REMOTE), we conducted sensitivity experiments, specifically analyzing the Lagrange multipliers $\beta_1, \beta_2$ in Eq. (1), the temperature factor $\tau$ in Eq. (5), and the loss weights $\lambda_1, \lambda_2, \lambda_3$ in Eq. (12). In the visualization, F1 scores for the UMRE, MORE, and MNRE datasets are color-coded in red, blue, and yellow, respectively. The results indicate that the model achieves optimal F1 scores when $\beta_1=\beta_2=1$ (Figure~\ref{fig:params}(a)), $\tau=0.175$ (Figure~\ref{fig:params}(b)), and $\lambda_1=1e-3, \lambda_2=1e-5, \lambda_3=1e-3$ (Figure~\ref{fig:params}(c)).

\begin{table}[t]
    \centering
    \caption{Detailed performance comparison on specific relation types from the UMRE dataset (F1-score). "Count" denotes the number of samples in the test set.}
    \label{tab:relations}
    \begin{tabular}{lccc}
        \toprule
        \textbf{Relation Type(Size)} & \textbf{REMOTE} & \textbf{Ours} \\
        \midrule
        /per/loc/place\_of\_governance(148) & 63.41 & 65.37 (\textcolor{red}{+1.96}) \\
        /per/misc/party(58) & 79.07 & 78.69 (\textcolor{blue}{-0.38}) \\
        /per/org/member\_of(650) & 60.05 & 61.67 (\textcolor{red}{+1.62}) \\
        /per/per/self(1070) & 89.85 & 89.71 (\textcolor{blue}{-0.14}) \\
        /per/misc/nationality(107) & 70.59 & 72.91 (\textcolor{red}{+2.32}) \\
        /loc/loc/self(180) & 90.17 & 92.05 (\textcolor{red}{+1.88}) \\
        /per/misc/present\_in(351) & 63.52 & 64.93 (\textcolor{red}{+1.41}) \\
        /per/loc/place\_of\_residence(189) & 40.58 & 43.17 (\textcolor{red}{+2.59}) \\
        /org/org/self(224) & 79.28 & 80.74 (\textcolor{red}{+1.46}) \\
        /misc/misc/self(172) & 79.75 & 82.97 (\textcolor{red}{+3.22}) \\
        /per/per/opponent(45) & 29.21 & 48.57 (\textcolor{red}{+19.36}) \\
        /per/loc/place\_of\_birth(96) & 41.11 & 37.97 (\textcolor{blue}{-3.41}) \\
        /per/per/partner(716) & 57.32 & 57.43 (\textcolor{red}{+0.11}) \\
        /per/org/opposed\_to(52) & 19.57 & 20.51 (\textcolor{red}{+0.94}) \\
        /loc/loc/contain(272) & 82.81 & 82.44 (\textcolor{blue}{-0.37}) \\
        /org/loc/locate\_at(123) & 52.38 & 52.21 (\textcolor{blue}{-0.17}) \\
        /per/misc/president(9) & 35.29 & 47.06 (\textcolor{red}{+11.77}) \\
        /misc/loc/held\_on(66) & 54.29 & 57.93 (\textcolor{red}{+3.64}) \\
        /per/org/leader\_of(43) & 51.11 & 60.67 (\textcolor{red}{+9.56}) \\
        /org/org/subsidiary(64) & 28.57 & 27.72 (\textcolor{blue}{-0.85}) \\
        /per/per/relatives(113) & 42.67 & 43.37 (\textcolor{red}{+0.70}) \\
        /per/misc/awarded(50) & 44.71 & 53.49 (\textcolor{red}{+8.78}) \\
        /misc/misc/part\_of(33) & 33.33 & 31.03 (\textcolor{blue}{-2.30}) \\
        /per/misc/race(2) & 0.00  & 66.67 (\textcolor{red}{+66.67}) \\
        /per/per/alumni(5) & 0.00  & 0.00 (0.00) \\
        /per/misc/religion(5) & 0.00  & 0.00 (0.00) \\
        /org/misc/present\_in(1) & 0.00  & 0.00 (0.00) \\
        \bottomrule
    \end{tabular}
\end{table}

\subsection{Analysis of Relation Types}

We evaluate UG-UMRE on the UMRE test set across relation types and sample sizes (Table~\ref{tab:relations}). The widespread \textcolor{red}{red} markers indicate consistent improvements over REMOTE, especially for long-tail and complex relations. For \textit{/per/per/opponent}, UG-UMRE improves F1 by \textbf{19.36\%}; for the extremely sparse \textit{/per/misc/race}, it reaches \textbf{66.67\%} F1 while REMOTE achieves zero. These gains suggest that probabilistic feature modeling and JAUA harmonize heterogeneous modalities and reveal semantics obscured by noise. Although extreme few-shot cases (count $\leq 5$) remain challenging, the results support the reliability of our uncertainty-guided approach.

\subsection{Computational Efficiency}

To quantify the deployment cost of UG-UMRE, we measure FLOPs, parameter count, peak GPU memory, and inference time on UMRE with batch size 16. As shown in Table~\ref{tab:efficiency}, integrating UG-UMRE into REMOTE adds 2.36M parameters (0.49\%), 3.7G FLOPs, 100 MiB peak memory, and 4.13 ms per sample, while improving F1 by 2.34 points. The same integration also produces a 1.92-point gain over FocalMRE with modest overhead. These measurements indicate that UDUA and JAUA can be added without changing the backbone hyperparameters, providing a practical accuracy--efficiency trade-off.

\begin{table}[t]
    \centering
    \caption{Computational efficiency on UMRE (batch\_size=16).}
    \label{tab:efficiency}
    \resizebox{\columnwidth}{!}{
    \begin{tabular}{lccccc}
        \toprule
        \textbf{Methods} & FLOPs (G) & Params (M) & Peak Mem (MiB) & Inference (ms) & F1 (\%) \\
        \midrule
        FocalMRE & 52.4 & 236.73 & 11924 & 64.98 & 63.57 \\
        + UG-UMRE & 56.8 & 239.09 & 12546 & 70.16 & 65.49 \\
        REMOTE & 106.5 & 483.92 & 15150 & 115.11 & 67.64 \\
        + UG-UMRE & 110.2 & 486.28 & 15250 & 119.24 & 69.98 \\
        \bottomrule
    \end{tabular}
    }
\end{table}
    
\subsection{Case Study}

\begin{figure}[t]
    \centering
\includegraphics[width=1.0\columnwidth]{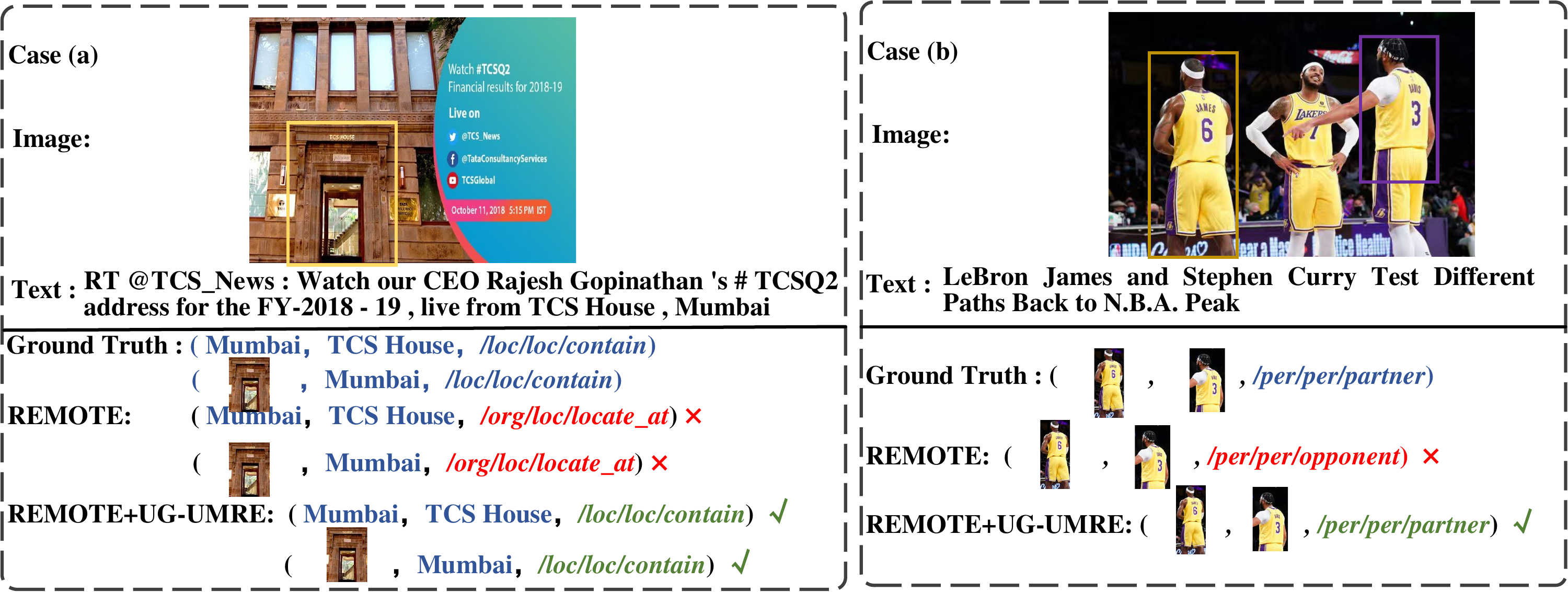}
\caption{Prediction Comparison for Two Test Samples on UMRE Dataset (\checkmark: Correct; $\times$: Incorrect).}
    \label{fig:case}
\end{figure}

As shown in Figure~\ref{fig:case}, we present a case study comparing our proposed REMOTE+UG-UMRE against the baseline REMOTE for cross-modal relational reasoning. 

In Case (a), our method correctly predicts the \textit{/loc/loc/contain} containment relation between textual entities and realizes precise cross-modal visual-text alignment, whereas REMOTE misclassifies the relation as \textit{/org/loc/locate\_at} due to distraction from organizational attributes in the text. In Case (b), our approach infers the correct \textit{/per/per/partner} teammate relation by aligning visual features of the bounded players with textual context, while REMOTE fails to leverage collaborative visual cues and erroneously outputs the \textit{/per/per/opponent} relation. 

This case validates the superiority of UG-UMRE in neutralizing semantic interference and bridging modal heterogeneity: the UDUA module filters visual noise to maintain semantic consistency, and the JAUA module enables fine-grained cross-modal alignment. With text as the semantic anchor, the denoised and calibrated visual features provide key contextual support, significantly boosting the accuracy of cross-modal relational reasoning.

\section{Conclusion}
We present UG-UMRE, an Uncertainty-Guided Unified Multimodal Relation Extraction network that addresses intra-modal noise and cross-modal distribution heterogeneity. Its UDUA module uses a variational information bottleneck to model features as Gaussian distributions, filtering aleatoric noise and improving semantic consistency. The JAUA module globally pre-calibrates cross-modal distributions, enabling coarse-grained alignment before fine-grained interaction. Experiments on three benchmarks demonstrate SOTA performance, favorable efficiency, and plug-and-play compatibility.

Current uncertainty estimates rely on holistic paired inputs and fixed loss weights, which may be less reliable for sparse relations. Future work will explore dynamic thresholds for few-shot and long-tail relations, as well as robust open-domain modeling under unknown noise, severe mismatch, and missing modalities.

\section*{Acknowledgments}

We sincerely thank the anonymous reviewers, Program Chairs, and Area Chairs for their constructive comments and careful handling of this submission. We also thank all participants and contributors involved in this work for their valuable support. This work was supported by the National Natural Science Foundation of China (No.62595731).

\bibliographystyle{ACM-Reference-Format}
\bibliography{sample-base}

@article{b1,
  author = "Xinkui Lin and Yongxiu Xu and Minghao Tang and Shilong Zhang and Hongbo Xu and Hao Xu and Yubin Wang",
  title = "{REMOTE}: {A} Unified Multimodal Relation Extraction Framework with Multilevel Optimal Transport and Mixture-of-Experts",
  journal = "arXiv preprint arXiv:2509.04844",
  year = "2025"
}

@inproceedings{b2,
  author = "Liang He and Hongke Wang and Zhen Wu and Jianbing Zhang and Xinyu Dai and Jiajun Chen",
  title = "Focus {\&} Gating: {A} Multimodal Approach for Unveiling Relations in Noisy Social Media",
  booktitle = "Proceedings of the MM",
  pages = "1379--1388",
  publisher = "ACM",
  year = "2024"
}

@inproceedings{b3,
  author = "Pengfei Wei and Zhaokang Huang and Hongjun Ouyang and Qintai Hu and Bi Zeng and Guang Feng",
  title = "{CGI-MRE}: {A} Comprehensive Genetic-Inspired Model For Multimodal Relation Extraction",
  booktitle = "Proceedings of the ICMR",
  pages = "524--532",
  publisher = "ACM",
  year = "2024"
}

@inproceedings{b4,
  author = "Xiyang Liu and Chunming Hu and Richong Zhang and Kai Sun and Samuel Mensah and Yongyi Mao",
  title = "Multimodal Relation Extraction via a Mixture of Hierarchical Visual Context Learners",
  booktitle = "Proceedings of the WWW",
  pages = "4283--4294",
  publisher = "ACM",
  year = "2024"
}

@article{b5,
  author = "Shiyao Cui and Jiangxia Cao and Xin Cong and Jiawei Sheng and Quangang Li and Tingwen Liu and Jinqiao Shi",
  title = "Enhancing Multimodal Entity and Relation Extraction With Variational Information Bottleneck",
  journal = "{IEEE/ACM} Transactions on Audio, Speech, and Language Processing",
  volume = "32",
  pages = "1274--1285",
  year = "2024"
}

@inproceedings{b6,
  author = "Liang He and Hongke Wang and Yongchang Cao and Zhen Wu and Jianbing Zhang and Xinyu Dai",
  title = "{MORE}: {A} Multimodal Object-Entity Relation Extraction Dataset with a Benchmark Evaluation",
  booktitle = "Proceedings of the MM",
  pages = "4564--4573",
  publisher = "ACM",
  year = "2023"
}

@inproceedings{b7,
  author = "Shengqiong Wu and Hao Fei and Yixin Cao and Lidong Bing and Tat{-}Seng Chua",
  title = "Information Screening whilst Exploiting! Multimodal Relation Extraction with Feature Denoising and Multimodal Topic Modeling",
  booktitle = "Proceedings of the ACL",
  pages = "14734--14751",
  publisher = "Association for Computational Linguistics",
  year = "2023"
}

@inproceedings{b8,
  author = "Lei Li and Xiang Chen and Shuofei Qiao and Feiyu Xiong and Huajun Chen and Ningyu Zhang",
  title = "On Analyzing the Role of Image for Visual-Enhanced Relation Extraction (Student Abstract)",
  booktitle = "Proceedings of the AAAI",
  pages = "16254--16255",
  publisher = "AAAI Press",
  year = "2023"
}

@inproceedings{b9,
  author = "Xiang Chen and Ningyu Zhang and Lei Li and Shumin Deng and Chuanqi Tan and Changliang Xu and Fei Huang and Luo Si and Huajun Chen",
  title = "Hybrid Transformer with Multi-level Fusion for Multimodal Knowledge Graph Completion",
  booktitle = "Proceedings of the SIGIR",
  pages = "904--915",
  publisher = "ACM",
  year = "2022"
}

@inproceedings{b10,
  author = "Changmeng Zheng and Junhao Feng and Yi Cai and Xiaoyong Wei and Qing Li",
  title = "Rethinking Multimodal Entity and Relation Extraction from a Translation Point of View",
  booktitle = "Proceedings of the ACL",
  pages = "6810--6824",
  publisher = "Association for Computational Linguistics",
  year = "2023"
}

@article{b11,
  author = "Wenti Huang and Jiayi Chen and Junjie Li and Yiyu Mao and Ningyi Mao",
  title = "{ES-MRE}: Evidence subgraph enhanced reasoning for multimodal relation extraction",
  journal = "Knowledge-Based Systems",
  volume = "325",
  pages = "113770",
  year = "2025"
}

@article{b12,
  author = "Xinyu He and Shixin Li and Yuning Zhang and Binhe Li and Sifan Xu and Yuqing Zhou",
  title = "The more quality information the better: Hierarchical generation of multi-evidence alignment and fusion model for multimodal entity and relation extraction",
  journal = "Information Processing {\&} Management",
  volume = "62",
  number = "1",
  pages = "103875",
  year = "2025"
}

@inproceedings{b13,
  author = "Zefan Zhang and Weiqi Zhang and Yanhui Li and Tian Bai",
  title = "Caption-Aware Multimodal Relation Extraction with Mutual Information Maximization",
  booktitle = "Proceedings of the MM",
  pages = "1148--1157",
  publisher = "ACM",
  year = "2024"
}

@inproceedings{b14,
  author = "Changmeng Zheng and Zhiwei Wu and Junhao Feng and Ze Fu and Yi Cai",
  title = "{MNRE}: {A} Challenge Multimodal Dataset for Neural Relation Extraction with Visual Evidence in Social Media Posts",
  booktitle = "Proceedings of the ICME",
  pages = "1--6",
  publisher = "IEEE",
  year = "2021"
}

@inproceedings{b15,
  author = "Zixian Gao and Xun Jiang and Xing Xu and Fumin Shen and Yujie Li and Heng Tao Shen",
  title = "Embracing Unimodal Aleatoric Uncertainty for Robust Multimodal Fusion",
  booktitle = "Proceedings of the CVPR",
  pages = "26866--26875",
  publisher = "IEEE",
  year = "2024"
}

@inproceedings{b16,
  author = "Yatai Ji and Junjie Wang and Yuan Gong and Lin Zhang and Yanru Zhu and Hongfa Wang and Jiaxing Zhang and Tetsuya Sakai and Yujiu Yang",
  title = "{MAP}: Multimodal Uncertainty-Aware Vision-Language Pre-training Model",
  booktitle = "Proceedings of the CVPR",
  pages = "23262--23271",
  publisher = "IEEE",
  year = "2023"
}

@inproceedings{b17,
  author = "Jielong Tang and Yang Yang and Jianxing Yu and Zhen-Xing Wang and Haoyuan Liang and Liang Yao and Jian Yin",
  title = "{U}n{C}o: Uncertainty-Driven Collaborative Framework of Large and Small Models for Grounded Multimodal {NER}",
  booktitle = "Proceedings of the EMNLP",
  pages = "7633--7651",
  publisher = "Association for Computational Linguistics",
  year = "2025"
}

@inproceedings{b18,
  author = "Junyu Gao and Mengyuan Chen and Changsheng Xu",
  title = "Collecting Cross-Modal Presence-Absence Evidence for Weakly-Supervised Audio-Visual Event Perception",
  booktitle = "Proceedings of the CVPR",
  pages = "18827--18836",
  publisher = "IEEE",
  year = "2023"
}

@inproceedings{b19,
  author = "Zixian Gao and Disen Hu and Xun Jiang and Huimin Lu and Heng Tao Shen and Xing Xu",
  title = "Enhanced Experts with Uncertainty-Aware Routing for Multimodal Sentiment Analysis",
  booktitle = "Proceedings of the MM",
  pages = "9650--9659",
  publisher = "ACM",
  year = "2024"
}

@inproceedings{b20,
  author = "Naftali Tishby and Noga Zaslavsky",
  title = "Deep learning and the information bottleneck principle",
  booktitle = "Proceedings of the ITW",
  pages = "1--5",
  publisher = "IEEE",
  year = "2015"
}

@inproceedings{b21,
  author = "Alexander A. Alemi and Ian Fischer and Joshua V. Dillon and Kevin Murphy",
  title = "Deep Variational Information Bottleneck",
  booktitle = "Proceedings of the ICLR",
  publisher = "OpenReview.net",
  year = "2017"
}

@article{b22,
  author = "Shuai Bai and Keqin Chen and Xuejing Liu and Jialin Wang and Wenbin Ge and Sibo Song and Kai Dang and Peng Wang and Shijie Wang and Jun Tang and Humen Zhong and Yuanzhi Zhu and Ming{-}Hsuan Yang and Zhaohai Li and Jianqiang Wan and Pengfei Wang and Wei Ding and Zheren Fu and Yiheng Xu and Jiabo Ye and Xi Zhang and Tianbao Xie and Zesen Cheng and Hang Zhang and Zhibo Yang and Haiyang Xu and Junyang Lin",
  title = "{Qwen2.5-VL} Technical Report",
  journal = "arXiv preprint arXiv:2502.13923",
  year = "2025"
}

@inproceedings{b23,
  author = "Jacob Devlin and Ming{-}Wei Chang and Kenton Lee and Kristina Toutanova",
  title = "{BERT}: Pre-training of Deep Bidirectional Transformers for Language Understanding",
  booktitle = "Proceedings of the NAACL",
  pages = "4171--4186",
  publisher = "Association for Computational Linguistics",
  year = "2019"
}

@inproceedings{b24,
  author = "Alexey Dosovitskiy and Lucas Beyer and Alexander Kolesnikov and Dirk Weissenborn and Xiaohua Zhai and Thomas Unterthiner and Mostafa Dehghani and Matthias Minderer and Georg Heigold and Sylvain Gelly and Jakob Uszkoreit and Neil Houlsby",
  title = "An Image is Worth 16x16 Words: Transformers for Image Recognition at Scale",
  booktitle = "Proceedings of the ICLR",
  publisher = "OpenReview.net",
  year = "2021"
}

@inproceedings{b25,
  author = "Lihe Yang and Bingyi Kang and Zilong Huang and Zhen Zhao and Xiaogang Xu and Jiashi Feng and Hengshuang Zhao",
  title = "Depth Anything {V2}",
  booktitle = "Proceedings of the NeurIPS",
  year = "2024"
}

@inproceedings{b26,
  author = "Haotian Liu and Chunyuan Li and Qingyang Wu and Yong Jae Lee",
  title = "Visual Instruction Tuning",
  booktitle = "Proceedings of the NeurIPS",
  year = "2023"
}

@inproceedings{b27,
  author = "Yaodong Yu and Tianzhe Chu and Shengbang Tong and Ziyang Wu and Druv Pai and Sam Buchanan and Yi Ma",
  title = "Emergence of Segmentation with Minimalistic White-Box Transformers",
  booktitle = "Proceedings of the CPAL",
  pages = "72--93",
  publisher = "PMLR",
  year = "2024"
}

@article{b28,
  author = "Abdelwahed Khamis and Russell Tsuchida and Mohamed Tarek and Vivien Rolland and Lars Petersson",
  title = "Scalable Optimal Transport Methods in Machine Learning: {A} Contemporary Survey",
  journal = "IEEE Transactions on Pattern Analysis and Machine Intelligence",
  pages = "1--20",
  year = "2024"
}

@article{b29,
  author = "Sina Moradi",
  title = "A Survey on Algorithmic Developments in Optimal Transport Problem with Applications",
  journal = "arXiv preprint arXiv:2501.06247",
  year = "2025"
}

@inproceedings{b30,
  author = "Marco Cuturi",
  title = "Sinkhorn Distances: Lightspeed Computation of Optimal Transport",
  booktitle = "Proceedings of the NeurIPS",
  pages = "2292--2300",
  year = "2013"
}

@inproceedings{b31,
  author = "Diederik P. Kingma and Max Welling",
  title = "Auto-Encoding Variational Bayes",
  booktitle = "Proceedings of the ICLR",
  year = "2014"
}

@inproceedings{b33,
  author = "Ilya Loshchilov and Frank Hutter",
  title = "Decoupled Weight Decay Regularization",
  booktitle = "Proceedings of the ICLR",
  publisher = "OpenReview.net",
  year = "2019"
}

@inproceedings{b34,
  author = "Changmeng Zheng and Junhao Feng and Ze Fu and Yi Cai and Qing Li and Tao Wang",
  title = "Multimodal Relation Extraction with Efficient Graph Alignment",
  booktitle = "Proceedings of the MM",
  pages = "5298--5306",
  publisher = "ACM",
  year = "2021"
}

@article{b35,
  author = "Peng Wang and Shuai Bai and Sinan Tan and Shijie Wang and Zhihao Fan and Jinze Bai and Keqin Chen and Xuejing Liu and Jialin Wang and Wenbin Ge and Yang Fan and Kai Dang and Mengfei Du and Xuancheng Ren and Rui Men and Dayiheng Liu and Chang Zhou and Jingren Zhou and Junyang Lin",
  title = "{Qwen2-VL}: Enhancing Vision-Language Model's Perception of the World at Any Resolution",
  journal = "arXiv preprint arXiv:2409.12191",
  year = "2024"
}

@article{b36,
  author = "{Llama Team}",
  title = "The {Llama} 3 Herd of Models",
  journal = "arXiv preprint arXiv:2407.21783",
  year = "2024"
}

@inproceedings{b37,
  author = "Robin Rombach and Andreas Blattmann and Dominik Lorenz and Patrick Esser and Bj{\"{o}}rn Ommer",
  title = "High-Resolution Image Synthesis with Latent Diffusion Models",
  booktitle = "Proceedings of the CVPR",
  pages = "10674--10685",
  publisher = "IEEE",
  year = "2022"
}

@inproceedings{b38,
  author = "Geoffrey E. Hinton and Sam T. Roweis",
  title = "Stochastic Neighbor Embedding",
  booktitle = "Proceedings of the NeurIPS",
  pages = "833--840",
  publisher = "MIT Press",
  year = "2002"
}

@inproceedings{b39,
  author = "Li Yuan and Yi Cai and Xudong Shen and Qing Li and Qingbao Huang and Zikun Deng and Tao Wang",
  title = "Collaborative Multi-{LoRA} Experts with Achievement-based Multi-Tasks Loss for Unified Multimodal Information Extraction",
  booktitle = "Proceedings of the IJCAI",
  pages = "6940--6948",
  year = "2025"
}

@inproceedings{b40,
  author = "Lin Sun and Kai Zhang and Qingyuan Li and Renze Lou",
  title = "{UMIE}: Unified Multimodal Information Extraction with Instruction Tuning",
  booktitle = "Proceedings of the AAAI",
  pages = "19062--19070",
  publisher = "AAAI Press",
  year = "2024"
}

@inproceedings{b42,
  author = "Zhiqiang Kou and Jing Wang and Jiawei Tang and Yuheng Jia and Boyu Shi and Xin Geng",
  title = "Exploiting Multi-Label Correlation in Label Distribution Learning",
  booktitle = "Proceedings of the IJCAI",
  pages = "4326--4334",
  year = "2024"
}

@inproceedings{b43,
  author = "Zhiqiang Kou and Si Qin and Hailin Wang and Jing Wang and Mingkun Xie and Shuo Chen and Yuheng Jia and Tongliang Liu and Masashi Sugiyama and Xin Geng",
  title = "Label Distribution Learning with Biased Annotations Assisted by Multi-Label Learning",
  booktitle = "Proceedings of the IJCAI",
  year = "2025"
}

@inproceedings{b44,
  author = "Zhiqiang Kou and Yucheng Xie and Hailin Wang and Jing Wang and Mingkun Xie and Shuo Chen and Yuheng Jia and Tongliang Liu and Xin Geng",
  title = "{RankMatch}: {A} Novel Approach to Semi-Supervised Label Distribution Learning Leveraging Rank Correlation between Labels",
  booktitle = "Proceedings of the NeurIPS",
  year = "2025"
}

@inproceedings{b45,
  author = "Zhiqiang Kou and Junxiang Wu and Wenke Huang and Wenwen He and Ming{-}Kun Xie and Changwei Wang and Yuheng Jia and Di Jiang and Yang Liu and Xin Geng and Qiang Yang",
  title = "{FedHarmony}: Harmonizing Heterogeneous Label Correlations in Federated Multi-Label Learning",
  booktitle = "Proceedings of the CVPR",
  publisher = "IEEE",
  year = "2026"
}

@article{b46,
  author = "Zhiqiang Kou and Junyang Chen and Xin{-}Qiang Cai and Xiaobo Xia and Ming{-}Kun Xie and Dong{-}Dong Wu and Biao Liu and Yuheng Jia and Xin Geng and Masashi Sugiyama and Tat{-}Seng Chua",
  title = "Positive-Unlabeled Reinforcement Learning Distillation for On-Premise Small Models",
  journal = "arXiv preprint arXiv:2601.20687",
  year = "2026"
}

@article{b47,
  author = "Junxiang Wu and Zhiqiang Kou and Hongwei Zeng and Wenke Huang and Biao Liu and Hanlin Gu and Yuheng Jia and Di Jiang and Yang Liu and Xin Geng",
  title = "Trustworthy Federated Label Distribution Learning under Annotation Quality Disparity",
  journal = "arXiv preprint arXiv:2605.04827",
  year = "2026"
}

@article{b48,
  author = "Zhiqiang Kou and Jing Wang and Yuheng Jia and Biao Liu and Xin Geng",
  title = "Instance-Dependent Inaccurate Label Distribution Learning",
  journal = "IEEE Transactions on Neural Networks and Learning Systems",
  volume = "36",
  number = "1",
  pages = "1425--1437",
  doi = "10.1109/TNNLS.2023.3329870",
  year = "2025"
}

@inproceedings{b49,
  author = "Bo Kong and Shengquan Liu and Liang He and Liruizhi Jia and Yi Liang",
  title = "{CSMA-CNER}: Multi-modal Chinese {NER} Task with Cross- and Self-Modality Attention",
  booktitle = "Proceedings of the ICME",
  pages = "1--6",
  publisher = "IEEE",
  doi = "10.1109/ICME57554.2024.10688285",
  year = "2024"
}

@article{b50,
  author = "Bo Kong and Shengquan Liu and Liruizhi Jia and Yi Liang and Dongfang Han and Xu Zhang",
  title = "{MINIGE-MNER}: {A} Multi-Stage Interaction Network Inspired by Gene Editing for Multimodal Named Entity Recognition",
  journal = "Neural Networks",
  volume = "194",
  pages = "108106",
  doi = "10.1016/j.neunet.2025.108106",
  year = "2026"
}

@inproceedings{b51,
  author = "Liruizhi Jia and Shengquan Liu and Bo Kong and Yuan Liu",
  title = "{REIA}: Entity Relation Extraction Based on Interaction Policy and Data Augmentation",
  booktitle = "Proceedings of the IJCNN",
  pages = "1--8",
  publisher = "IEEE",
  doi = "10.1109/IJCNN64981.2025.11227659",
  year = "2025"
}

@article{b52,
  author = "Changpeng Zhao and Dongfang Han and Zicheng Zuo and Turdi Tohti",
  title = "{KGDB-DDI}: Knowledge Graph-Based Drug Background Data Fusion Model for Drug-Drug Interaction Prediction",
  journal = "Artificial Intelligence in Medicine",
  volume = "168",
  pages = "103225",
  doi = "10.1016/j.artmed.2025.103225",
  year = "2025"
}

@article{b53,
  author = "Yi Liang and Turdi Tohti and Wenpeng Hu and Tianwei Yan and Shaohuang Wang and Askar Hamdulla",
  title = "{LLaMA-MoT}: {A} Cost-Effective Framework for Visual-Linguistic Instruction Tuning Based on Multi-Head Adapters and Chain-of-Thought",
  journal = "Expert Systems with Applications",
  volume = "297",
  pages = "129250",
  doi = "10.1016/j.eswa.2025.129250",
  year = "2026"
}

\end{document}